\newif\ifarxivversion
\arxivversiontrue% Switch here for the arxiv version
\ifarxivversion
\documentclass[10pt,cleanfoot]{asme2ej}
\else 
\documentclass[10pt]{asme2ej}
\fi

\usepackage{epsfig} %% for loading postscript figures
\usepackage{booktabs}
\usepackage[hidelinks]{hyperref}
\usepackage{caption}
\usepackage{subcaption}
\usepackage{tabularx}
\usepackage[dvipsnames]{xcolor}
\usepackage{soul,color} % hl{} macro

\title{What's in a Name? Evaluating Assembly-Part Semantic Knowledge in Language Models through User-Provided Names in CAD Files}

\author{Peter Meltzer
    \affiliation{
    Autodesk Research \\
    London, UK
    }
}

\author{Joseph G. Lambourne
    \affiliation{
    Autodesk Research \\
    London, UK
    }
}

\author{Daniele Grandi
    \affiliation{
    Autodesk Research \\
    San Fransciso, USA
    }
}

\newcommand{\etal}{\textit{et al.\ }}

\newcommand\extrafootertext[1]{%
    \bgroup
    \renewcommand\thefootnote{\fnsymbol{footnote}}%
    \renewcommand\thempfootnote{\fnsymbol{mpfootnote}}%
    \footnotetext[0]{#1}%
    \egroup
}

\begin{document}

\maketitle

%%%%%%%%%%%%%%%%%%%%%%%%%%%%%%%%%%%%%%%%%%%%%%%%%%%%%%%%%%%%%%%%%%%%%%
\begin{abstract}
{\it 
Semantic knowledge of part-part and part-whole relationships in assemblies is useful for a variety of tasks from searching design repositories to the construction of engineering knowledge bases. In this work we propose that the natural language names designers use in Computer Aided Design (CAD) software are a valuable source of such knowledge, and that Large Language Models (LLMs) contain useful domain-specific information for working with this data as well as other CAD and engineering-related tasks.

In particular we extract and clean a large corpus of natural language part, feature and document names and use this to quantitatively demonstrate that a pre-trained language model can outperform numerous benchmarks on three self-supervised tasks, without ever having seen this data before. Moreover, we show that fine-tuning on the text data corpus further boosts the performance on all tasks, thus demonstrating the value of the text data which until now has been largely ignored. We also identify key limitations to using LLMs with text data alone, and our findings provide a strong motivation for further work into multi-modal text-geometry models.

To aid and encourage further work in this area we make all our data and code publicly available.

}
\end{abstract}

%%%%%%%%%%%%%%%%%%%%%%%%%%%%%%%%%%%%%%%%%%%%%%%%%%%%%%%%%%%%%%%%%%%%%%
%% Note:  On the submission form there were a set of standard choices

\section{Introduction}

\label{sec:intro}

In the mechanical engineering domain, natural language is used by designers and engineers throughout the design process: to express design requirements, document design intent, and communicate ideas and solutions to others in the complex network of people working together to create a single product. Computer-Aided Design (CAD) software is often used to document and communicate design decisions, where designers create assemblies of parts, and use natural language to name each part and the assembly itself for documentation and collaboration purposes.  This text often contains important semantic information about the individual parts and/or the roles they play in an assembly.  Additionally, by grouping the text associated with the parts in an assembly together, there is the potential for whole-part relationships to be recovered.  

While the geometry of parts and their relationships in assemblies have been extensively studied \cite{Bespalov2005,kwon2022, korbi2022cad, jones2021automate, willis2022joinable}, 
and various data mining and machine learning techniques have been used to build knowledge bases relevant to design \cite{Shi2017, sarica2020, Feng2020}, the natural language inside CAD models has been largely ignored \cite{Schinko2017}.  Recently, large language models (LLMs) have revolutionized the field of natural language processing (NLP), becoming the standard tool used for machine translation \cite{tan2020neural} and showing impressive performance on standard tasks like text summarizing \cite{raffel2020exploring} and question answering  in a few shot setting \cite{brown2020language}.  

As these models are trained on a broad range of human knowledge \cite{wiki, Zhu_2015_ICCV, Mehmood2017CommonCrawl}, they have some exposure to the domain-specific vocabulary and mechanical engineering concepts useful for understanding mechanical CAD data.  In this work, we evaluate their effectiveness at solving three problems that designers currently face when querying CAD content libraries and design repositories or managing the designs in product lifecycle management (PLM) systems.

The use of CAD content libraries greatly speeds up the design process, allowing engineers and designers to import models of standard parts and automatically add the part numbers to a bill of materials.  However searching these libraries for parts is often a tedious process.  One way the process can be accelerated is for the content library to provide recommendations for additional parts which may be suitable to add to the same assembly akin to a ``recently downloaded" or ``frequently purchased together" part.  To facilitate this, we investigate the ability of large language models to predict whether two parts are commonly found in the same assembly.

CAD content libraries may also benefit from having a better understanding of the model a designer is currently working on, allowing more relevant recommendations and better targeted search functionality.   Utilizing the text in CAD models to help provide this contextual information is an appealing solution, as the data can be efficiently extracted from the CAD software without computationally expensive interrogation of the model geometry and rapidly transmitted to a content provision server.  To validate the ability of large language models to utilize contextual information from CAD part names to improve recommendations, we analyze the accuracy with which they can predict the name of a randomly selected part when the other non-default part names in the assembly are provided.  In addition to content recommendations, this kind of prediction could be used to identify opportunities for the re-use of parts from a PLM system during the modeling process.

A final problem, which large language models may help address, is to be able to identify the type of object modelled by a given assembly from a list of its part names.  The ability to cluster and group assemblies together without the need for expensive geometric analysis would allow automatic categorization of assemblies in a PLM system.   Additionally, CAD software could use this kind of information to identify the kinds of parts being modelled and show more relevant options on quick access toolbars and provide content-aware help and documentation.  As there are no large datasets of CAD assemblies classified by object category, we instead investigate the ability of large language models to predict the user-defined names of OnShape documents.   As around 20\% of OnShape document names consist of a few words describing the assembly as a whole (e.g. ``Coffee Mug", ``Mechanical Pencil"), the ability to predict assembly names indicates the model could be used to group similar assemblies and classify them using few-shot learning techniques.

By evaluating the performance of a pre-trained language model at these three tasks, we also gain an understanding of the amount of mechanical engineering information which is inherited from its initial training and can study the amount by which performance can be increased by fine-tuning.

Our key contributions are:
\begin{enumerate}
    \item We extract and clean a natural language corpus built from all the non-default text strings in the ABC dataset \cite{koch2019abc}.   This includes the names of parts and modeling features along with the name of the document which contained them.
    \item We quantitatively evaluate a pre-trained DistilBERT language model on three self-supervised tasks.  We demonstrate that, without fine-tuning, the language model has superior performance over numerous benchmarks including TechNet \cite{sarica2020}, which was specifically designed to extract mechanical engineering knowledge from patent data.
    \item We show that fine-tuning the language model allows further improvements in performance.  The model is able to learn new relationships between part numbers and other domain-specific vocabulary, providing a pathway to better understand part assembly relationships by training on larger datasets.
\end{enumerate}

The source code and data for this project are available at \url{https://github.com/AutodeskAILab/WhatsInAName}.

\section{Related work}
\subsection{Transformers}
Since their introduction by Vaswani \etal \cite{vaswani2017attention}, transformer models have become the predominant network architecture used for a wide range of NLP tasks such as machine translation \cite{tan2020neural}, question answering \cite{ZHANG2021205} and text summarization \cite{raffel2020exploring}.  Given an array of tokens, derived from words in the input text, they are trained either to predict the next token in the sequence \cite{Radford2018} or the values of certain masked tokens \cite{devlin2019}.  
The key innovation in the transformer architecture is the self-attention mechanism, which allows the model to learn to focus on certain combinations of tokens.  Scaling up both the size of the transformers and the corpus they are trained on, has allowed good performance on many NLP tasks with minimal fine-tuning \cite{brown2020language}.  
The DistilBERT model \cite{sanh2020} is especially designed for transfer learning and fine-tuning on new tasks, and is initially trained to mimic the outputs of the full-size version of BERT \cite{devlin2019}, a LLM that has been successfully leveraged in both academia and industry \cite{rogers2020}, while requiring much less memory and being faster to compute. The data for pre-training includes English Wikipedia \cite{wiki} and BookCorpus \cite{Zhu_2015_ICCV}.

As the parts in CAD models are not canonically ordered, it is useful to work with their names as an unordered set.  The Set Transformer \cite{lee2019} has the advantage of permutation invariance with respect to the input tokens.   It also avoids the quadratic complexity which is usually present in the self-attention mechanism, allowing long lists of data to be consumed.  

Beyond conventional NLP settings alone, advances in language models have also enabled great progress in multi-modal tasks such as zero-shot image classification \cite{radford2021} and text-to-image generation \cite{rombach2022}. As of yet, there has been no work done on the analogous text-CAD applications.

\subsection{CAD search, retrieval and clustering}
 
One area where the text from CAD models has been utilized is for search and retrieval, however most published work in this area focuses on geometric similarity \cite{Schinko2017, Tangelder2004}.  Of the 50 papers analysed by Schinko \etal \cite{Schinko2017}, only 6 of them utilize metadata of any kind. Fisher \etal \cite{Fisher2010} incorporate a keyword search into a 3D model finding algorithm.  The text is derived from the CAD filename and scene graph nodes, and matching is done based on the intersection over union of the sets of words in the query.  Funkhouser \etal \cite{Funkhouser2003} crawled the web looking for VRML models and derived text annotations from the filenames and nearby text on the websites.   The text is cleaned by removing common words using the SMART document retrieval system \cite{Salton1991} and then matches were made using the TF-IDF/Rocchio method \cite{rocchio71relevance}.  
Yi \etal \cite{Yi2017} utilized part names in the ShapeNet scene graph \cite{chang2015shapenet} to learn shape hierarchies using an Expectation-Maximization (EM) algorithm, however they manually pre-process the noisy user-provided text into
a tag dictionary rather than feeding it directly to a machine learning model.  None of these works evaluate the utility of the raw text data for understanding parts that commonly occur in the same assembly, or for discovering part-whole relationships. 

\subsection{Multi-modal models}
Another area where text has been used is in the creation of multi-modal models.  These models embed both text and geometry information into a joint embedding space.    Chen \etal \cite{chen2018text2shape} collected natural language descriptions for chairs and tables in the ShapeNet dataset.   The text is encoded using a character level CNN and RNN (GRU) and the geometry encoded with a 3D CNN.  The text and geometry embeddings were built into a joint embedding space which could be used for search and generation using a Wasserstein GAN framework. Han \etal \cite{Han2019} used the same text data with sequences of multi-view images of the 3D models.  An RNN encoder is used to encode the data for each modality, and two RNN decoders jointly reconstruct the modality itself and predict the data for the counterpart modality.  More recently a number of works have used pre-trained CLIP models \cite{radford2021} for text-to-image generation \cite{rombach2022, ramesh2021zero, dalle}, text-to-3D generation \cite{sanghi2022clip, khalid2022clip, sanghi2022textcraft}  and text based search \cite{Schlachter2022}.

\subsection{Engineering focused knowledge bases}
Engineering knowledge bases, or design repositories, are databases that contain product design knowledge useful to design engineers \cite{Szykman2000,Bohm2004,Bohm2008}. Over the years, various schemas and taxonomies have been proposed to organize design knowledge\cite{Phelan2014,Bharadwaj2019,Kurtoglu2005,Hirtz2002,Cheong2011}; however, despite efforts to streamline the process of adding data to design repositories \cite{Ferrero2020a}, adoption in research and industry has been limited, due to the resource commitment necessary to expand these knowledge bases, and the costly manual human input required. 

Recent efforts have facilitated the creation of engineering knowledge bases via unsupervised knowledge-mining methods.
Shi \cite{Shi2017} constructed a semantic network for design and engineering knowledge by text mining close to a million mechanical engineering papers from ScienceDirect \cite{sciencedirect}; resulting in an ontology with a higher retrieval rate for engineering concepts than generic knowledge bases like WordNet \cite{Miller1995}, ConceptNet \cite{Liu2004}, and NeLL \cite{carlson2010}.
Sarica \etal created the TechNet \cite{sarica2020} semantic network using the titles and abstracts of 5 million patents from the USPTO patent database, and used a word2vec model \cite{Mikolov2013word2vec} to create vector embeddings for 4 million entities. Technet embeddings have been used to encode the semantic names of parts in CAD models \cite{bian2022material}.

Our work differs from previous research in that we leverage the knowledge of parts and their relationships available in a pre-trained LLM, rather than seeking to extract and represent it explicitly using specialized corpora or manually constructed knowledge bases.

\section{Data}

A key contribution of this work is the extraction and publication of a cleaned natural language corpus of  text strings from the ABC dataset, which we primarily use for evaluation purposes. We now describe the methodology used and provide a snapshot of the types of strings included.

\begin{table}[b]
    \centering
    \caption{Sample document names grouped according to how much the names tell us about the parts included inside. Percentage figures show estimates of the proportions of each string type based on a manually labeled random sample of 250 document names.}
    \begin{tabular}{llll}
          \toprule
Clean Semantic & 20.8\% & Coffee Mug & Mechanical Pencil \\
                && OS kinematics & OS Chess \\
                && Torch Light For Bike & Yoke \\
                && Bottle & Concept Vehicle \\
                && Mounting Arm  & \\
                \midrule
Noisy Semantic  & 34.8\% & Lava Lamp 2 & Sample - Headphones \\
                && Dave's Handsome Mug & Sample - Bicycle Helmet \\
                \midrule
Ambiguous       & 44.4\% & Assem Test & Sample Document 2 \\
                && Left.x\_t & Part Performance test \\
                && Part Per. Test 2 & Fasteners\_onshape\_Support \\
                && Mark's First Document \\
          \bottomrule
    \end{tabular}
    \label{tab:example_names}
\end{table}

\subsection{Data Extraction}
The text data used in this work was extracted from the ABC dataset \cite{koch2019abc}, which contains CAD models created by users of the OnShape software.   OnShape models are organized into ``documents", which can contain multiple ``tabs" for parts, assemblies, and drawings.  The dataset contains metadata for each document, including a user-defined document name and any additional text the user has supplied as a description. 

In this work we make the assumption that all the solid bodies inside a given document are related in some way.   Designers can choose to model multiple parts in a single tab, or to design each part in a separate tab and then combine them into an assembly. 
While not all documents represent assemblies, we consider most documents to contain a similar combination of parts to those found in an assembly.

We extract the part names from the \texttt{MANIFOLD\_SOLID\_BREP} entities of the STEP files and aggregate these for all STEP files in a document.  If the same string appears more than once in a single STEP file, then we maintain the largest multiplicity this string has for any STEP file derived from the document.   This preserves the multiplicity of parts that commonly appear more than once in a design  (4 wheels on a car) while avoiding duplication due to multiple design revisions of parts, which are sometimes kept in separate tabs. 
In addition, we extract the names of all  modeling features from the ABC metadata files.   Many of the part names and feature names are the defaults automatically generated by the OnShape software.  For example, solids are often named ``Part $n$", where $n$ is some small integer.   The default feature names include the feature type and an integer, i.e.  ``Extrude $n$" or ``Revolve $n$".    These default names are removed.  While only 47\% of the solids in the entire ABC dataset contain non-default names, we find that in documents which contain at least one non-default part name, $77\%$ of solids' names are non-default.   

The ABC dataset contains many duplicate documents.   To remove duplicates we take all the strings extracted from a given document and sort them.  If this sorted list of strings is identical to any other in the dataset then the two documents are considered as a duplicate and one copy is removed.   We also remove the prefix ``Copy of" and suffix ``- Copy" from document names as these affixes appear to have been added automatically by OnShape when documents are cloned.

The full ABC dataset contains files from 456,811 OnShape documents.   Of these, 80,282 documents have at least one non-default part name or feature name after deduplication.   Inside these documents we have a total of 950,335 non-default part names and 562,339 non-default feature names.   39,613 documents contain two or more non-default part names. The dataset is divided into a $70\%$/$15\%$/$15\%$ train, validation, test split.

In all our experiments, we further pre-process the dataset by converting all strings to lowercase and replacing all underscores with whitespace to remove potential opportunities for the models to cheat by spotting common naming conventions between parts and names from the same documents. However, we preserve the original cases in the version which we share in order to enable their use in further research if required.

%%%%%%%%%%%%%%%%%%%%%%%%%
% Data Discussion
%%%%%%%%%%%%%%%%%%%%%%%%%
\subsection{Data Snapshot}

\begin{table*}[b]
    \centering
    \caption{Parts from a sample of 10 documents grouped according to the level and interpretability of the semantics included. The percentages indicate the estimated fraction of all part names in the dataset with strings of a similar type.   A further 20.4\% of the strings were ambiguous.}
    \begin{tabularx}{\linewidth}{p{3cm}rX}
    \toprule
Part Codes or \newline Materials? & 1.6\%& OSB-O-S-4, OSB-I-S-3, OSB-O-J-3, ram-krokev-j, OSB-I-J-4, OSB-O-S-1, OSB-O-J-2, ram-pozed, OSB-I-J-5, OSB-I-J-2, hotovo, OSB-I-S-4, OSB-I-S-1, ram-pozed-vym, OSB-O-J-5, deska-pozed, OSB-I-J-1, OSB-I-J-3, vrchol-O, OSB-O-0, ram-krokev-s, OSB-O-J-1, OSB-O-S-2, OSB-O-S-3, OSB-I-S-5, OSB-O-S-5, OSB-I-S-2, OSB-O-J-4 \\

Part Codes or \newline Naming Convention? &  15.6\% & 47065t604, 47065t217, 47065t258, 47065t601, 47065t261, 47065t254, 47065t845, i\_id6\_x\_t \\

Clean Natural \newline Language & 23.6\% & Bishop, Knight, Queen, Pawn, Castle, King \\

& & room, door, window \\

& & upright, cylinder, cylinder top, shaft, baseplate, crank, bearing, flywheel, con-rod, socket, wheel pin, handle, back plate, rod pin, handle shaft \\

Natural Language \newline with Ordinal \newline Numbers & 14.8\%& Bathtub, House Frame, 1st Floor, Toilet, Walls, Utility Room, Walls, Basement Bath, Plumbing, Sink, 1st Floor Ceiling, 2nd Floor, Basement Ceiling, 2nd Floor Stairs, Basement Stairs, Walls, Basement, \\

&& Upright 2, Big Gear Rod, Pie Holder of Doom, Hangar 1, Randomizer 1, Cross Support 2, Throwing Arm 1, Randomizer 2, Little Gear, Big Gear, Throwing Arm Rod, Randomizer 3, CAM, Randomizer 4, Propeller Rod, Side 2, LIttle Gear Rod, Upright Support 1, Hangar 2, Upright Support 2, Upright 1, Propeller, Side 1, Throwing Arm 2, Cross Support 1, Upright Support 3, \\

Natural Language \newline with Dimensions & 8.4\%&18x24x76 Cabinet, 6x24x76 Cabinet, 23x24 Shelf \\

Natural Language \newline with Abbreviations & 9.6\%&flat knob, REVERSE MBOX, domed knob, d.knob moldbox \\

Fasteners & 2.8\% & \_Slotted head cap screw ISO 4762 - M6x16, \_Hexagon socket set screw with flat point DIN 913 - M5x5\_2, \_Slotted head cap screw ISO 4762 - M6x16\_3,\_Hexagon socket set screw with flat point DIN 913 - M5x5\_1,  \_Hexagon socket set screw with flat point DIN 913 - M5x5\_3, \_Hexagon socket set screw with flat point DIN 913 - M5x5, \_Slotted head cap screw ISO 4762 - M6x16\_1, \_Slotted head cap screw ISO 4762 - M6x16\_2 \\

Non-English with \newline Special Characters& 3.6\% &\textbackslash\textbackslash X2\textbackslash\textbackslash 00E9\textbackslash\textbackslash X0\textbackslash\textbackslash querre, cale mur, vasque, cale meubles \\
    \bottomrule
    \end{tabularx}
    \label{tab:example_parts}
\end{table*}

We provide a small snapshot of the extracted text data here to illustrate some of the challenges it presents.
In \autoref{tab:example_names} we show a sample of 20 document names organised according to the semantic relationship between the name and the parts contained in the document. The ``Clean Semantic" names present the most ideal case, as we have a strong indication of the contents of the document. For ``Noisy Semantic" it is also clear, but the names contain noise which may affect automated processing or use in language models. The ``Ambiguous" names present a worst-case scenario since, with respect to mechanical engineering or assembly design, they tell us nothing about the contained parts.

To get a sense of the distribution of these categories in the rest of the dataset, we manually classified 250 randomly selected document names.  In cases where domain-specific vocabulary or product names were used, we searched the internet to establish if the object could be identified from the document name without ambiguity.   The percentage of the document names in each category are shown in the table.

\autoref{tab:example_parts} shows the parts from a sample of 11 documents. Again there is a varied level of semantics present. In some cases it is obvious what the assembly or collection of parts would be, i.e. a chess set or a house. There are also many cases of materials, i.e. Oriented-Strand-Board (OSB), and commercial part codes. However, sometimes it is unclear whether or not we have part codes or custom naming conventions, i.e. an internet search for \texttt{47065t604} does not show this to be part code in common usage.  Other noise such as dimensions may also weaken the signal in the semantics of a part, while the inclusion of many non-English languages could also present a challenge, as we do not have enough strings from each language to enable multi-lingual generalization.

Fasteners are one of the most common types of component in mechanical assemblies.  Using the database of fasteners which Autodesk Inventor provides as its content library, we are able to automatically detect fasteners in the dataset with types `BS', `KS', `DIN EN', `DIN', `ISO', `AS', `UNI', `IS', `ANSI' and `DIN EN ISO'.   Fasteners are considered to be identified without ambiguity if the fastener type and dimensions are matched in a part name.  For example, an ``ISO 4762 Cap screw " would need to match the substrings ``ISO"and ``4762" with optional whitespace hyphens and dimensions of the form ``M$n_1$x$n_2$" where $n_1$ and $n_2$ are integer values which are valid dimensions for that fastener type. We find that 471 documents contains at least one fastener and 5008 fasteners are identified in total without ambiguity.

To provide an estimate of the quantities of different kinds of part name strings in the dataset, we randomly select 250 part names and classify them into the categories in  \autoref{tab:example_parts}.  The middle column of \autoref{tab:example_parts} shows the estimated fraction of all unique strings in the dataset which fall into each category.  To further quantify the fraction of the text with some semantic content, we count the number of unique strings containing words which match nouns which ConceptNet has tagged as ``artifacts" \cite{Liu2004}.  We find $51\%$ of the part names contain at least one word matching this criteria.  This is approximately consistent with the fraction of strings in the ``Clean natural language", ``Natural language with ordinal numbers", ``Natural language with dimensions", ``Natural language with abbreviations" and ``Fasteners" categories.
\section{Method}

\subsection{Part and Document Name Representations}

To encode the strings in the part names we seek to leverage the prior knowledge in a pre-trained version of DistilBERT \cite{sanh2020}. Specifically, we use \texttt{distilbert-base-uncased} from Hugging Face \cite{huggingface}.

Since each part string can be formed from multiple words, and even each word can be formed from multiple tokens, we pool the outputs of the language model to form a single vector for each part/document name string. Our initial experiments indicated that taking the mean of the output tokens in the second to last layer to be most effective.

\subsection{Document Representations}

We have two key considerations to take into account when representing a whole document. First, there can be an arbitrary number of parts, and second, the parts could appear in any order. Thus, we opt for a Set Transformer \cite{lee2019} to encode the set of embeddings for each part, since the set transformer is invariant to the order of parts and can handle arbitrarily sized input sets.

\begin{figure*}[b]
    \centering
    \includegraphics[width=\linewidth]{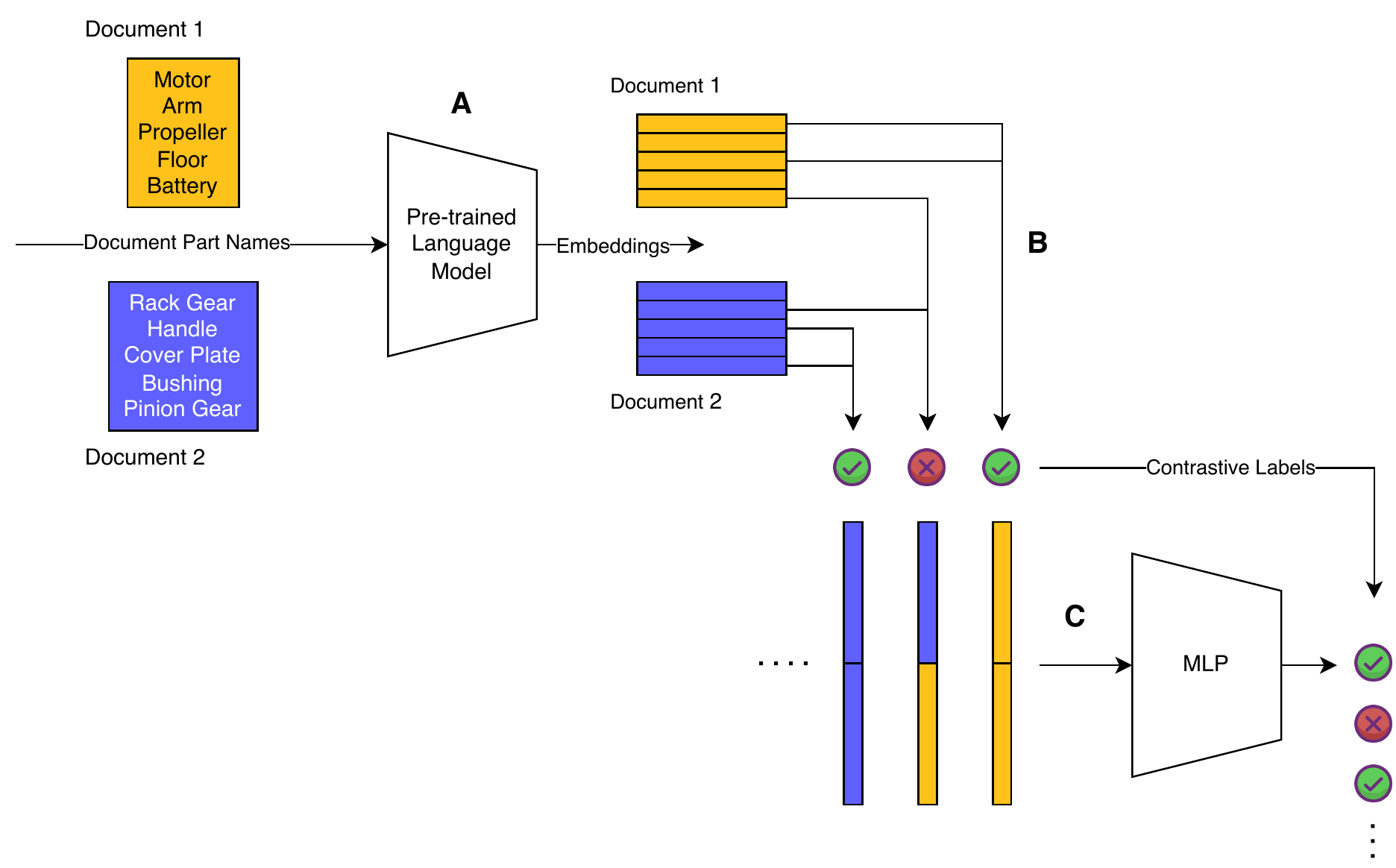}
    \caption{Architecture for training and evaluation of ``Two Parts" experiment. \textbf{A}: The parts from documents are embedded by the language model which is pre-trained and has all its weights fixed. \textbf{B}: Parts are sampled either from the same assembly to produce positive pairs, or from different assemblies to produce negative pairs. \textbf{C}: The labels generated by the sampling process, along with the positive/negative pairs, are used to train and evaluate a small MLP.}
    \label{fig:pairs}
\end{figure*}

\subsection{Self-Supervised Evaluation}

We design three self-supervised tasks to quantitatively evaluate the ability of LLMs to address the search and recommendation problems described in the introduction and our hypothesis that they contain useful knowledge relevant to mechanical engineering.  The performance at these tasks can then be compared against numerous baselines. The tasks are intended to evaluate each model's understanding of which parts commonly occur together in assemblies, and how easily the nature of the entire assembly can be understood from a list of its component parts. First, we consider whether the language model contains useful signals with which to tell if two parts are from the same document or not. Next, we consider how well a randomly selected part from a document can be identified given the remaining parts, and finally, we consider how well the document name can be identified from the document's parts. The design of these tasks thus enables quantitative comparisons between different language models and baselines despite the lack of ground truth labels in the dataset's original form.

We aim to evaluate how much information the DistilBERT language model knows about mechanical design, based only on its pre-training, and to measure how much fine-tuning will further improve performance.  To achieve this, the three self-supervised  tasks are used to evaluate both the base and fine-tuned DistilBERT models and the results are reported in \autoref{sec:results}.  In all experiments, fine-tuning of the language model is done using the training set only, using a standard LLM fine-tuning task (see \autoref{sec:fine-tuning}) rather than fine-tuning directly on each task. The training set is also used to train any baseline methods that do not have pre-trained embeddings available.

\subsubsection{Two Parts}
The ability to identify whether pairs of parts commonly co-occur in the same assembly is useful for CAD content recommender systems.  We frame  this task as a binary classification problem, in which a small Multi-Layer Perceptron (MLP) must predict whether or not two parts are taken from the same document. The input to the MLP is the concatenated LLM embeddings of either a positive or negative pair of parts, where positive pairs belong to the same document and negative pairs are from different documents, and the MLP is trained using a Binary Cross Entropy (BCE) loss. A high-level view of the architecture is shown in \autoref{fig:pairs}.

This experiment can also be used to assess the level of information about mechanical design present in a pre-trained LLM, as well as our baseline methods. As such, we opt for a small downstream model (the MLP), which contains only a single layer of 100 hidden units - this limited capacity ensures that the downstream network must rely most heavily on the signals already present in the language model embeddings to solve the task. All weights in the LLM are fixed during training of the MLP.

The labels are generated in a contrastive approach by sampling positive and negative pairs from the dataset. Positive pairs are formed of two parts taken from the same document, while negative pairs are sampled from different documents. To select the positive pairs, we consider only parts where the sets of tokens are distinct.  This excludes pairs with the same part names or pairs containing the same sub-words.  Here we use NLTK Wordpunkt tokens \cite{birdsteven2009} for each part. We then shuffle the pairwise correspondence to produce the negative pairs, and any resulting negative pairs which actually occur together somewhere in the dataset are removed. Since this results in fewer negative than positive pairs, we discard any additional positive pairs at random to balance the class distribution. Thus our result is a balanced binary classification dataset, where a random classifier would be expected to score 50\% accuracy.

Due to the small size of the evaluation model (the MLP), we further split the held-out test set into a subset for train, validation, and test with the same ratios as before. This gives us approximately 16.5K/3.5K/3.5K pairs for train, validation, and test respectively. This means the MLP is trained and evaluated on data that none of the language models have ever seen before, allowing us to compare against  language models fine-tuned using the original train data, and provides a good means to evaluate generalization.

\subsubsection{Missing Part}

\begin{figure*}[t]
    \centering
    \begin{subfigure}[b]{\linewidth}
        \includegraphics[width=\linewidth]{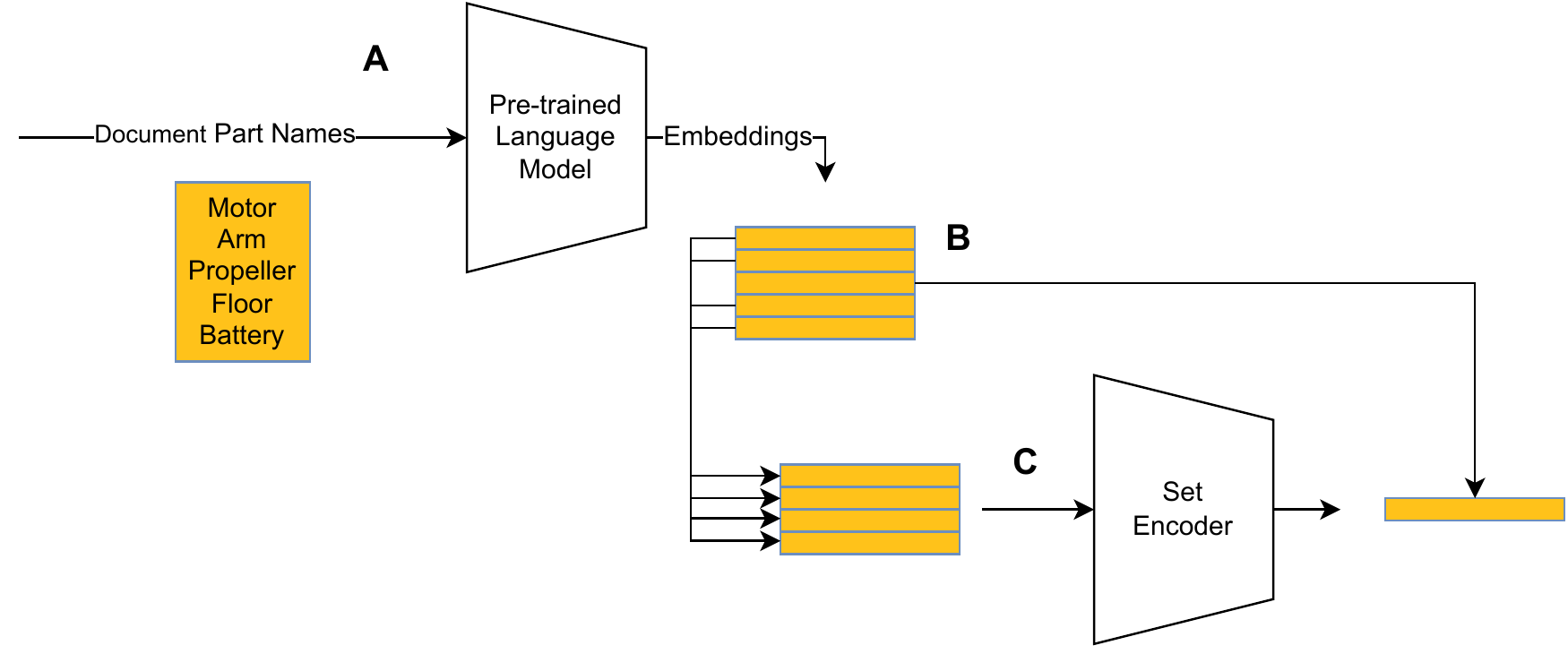}
        \caption{Training. \textbf{A}: The part strings for each document are embedded by the language model. \textbf{B}: One part is sampled at random and used as a target, the remaining parts are given as an input set. \textbf{C}: The set encoder is trained auto-regressively to predict the embedding of the missing part using an MSE loss. The language model is fixed during training.}
        \label{fig:missing_part_train}
    \end{subfigure}
    \vbox{}
    \begin{subfigure}[b]{\linewidth}
        \includegraphics[width=\linewidth]{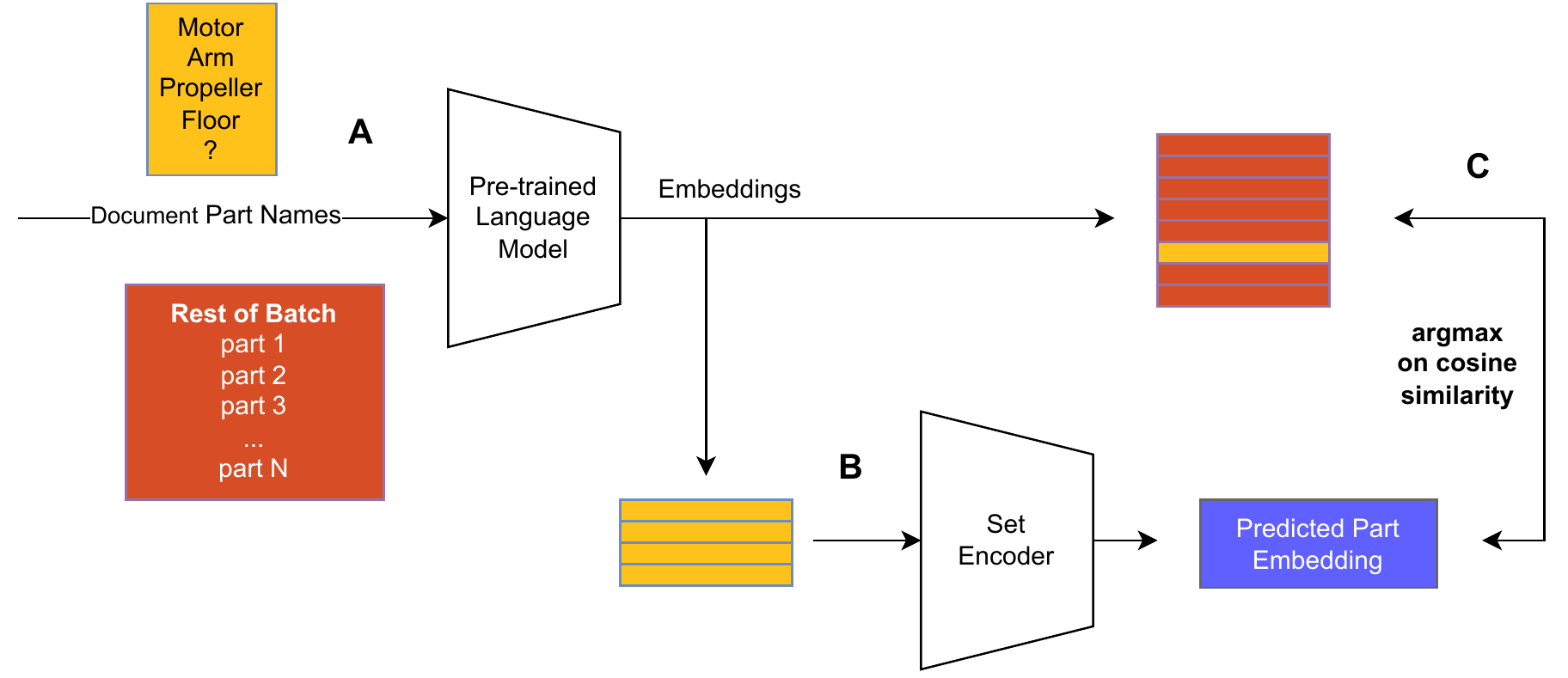}
        \caption{Inference/Evaluation. \textbf{A}: A batch containing part strings from multiple documents is embedded. \textbf{B}: For each document, a part embedding is removed at random and the set encoder predicts its embedding. \textbf{C}: The removed part embedding is mixed in with the part embeddings for the rest of the batch to form candidates, and the cosine similarity is used to select the closest candidate to the predicted part embedding.}
        \label{fig:missing_part_inference}
    \end{subfigure}
    \caption{Architecture for ``Missing Part" experiment.}
\end{figure*}

Next, we wish to evaluate our approach at predicting a missing part from a document, which is directly applicable to recommending a commercially available off-the-shelf part that a designer might want to insert into an assembly. For the documents from which text was extracted to create our dataset, 23\% of the part names were left as default and 77\% had user-provided names, thus by excluding the default part names we have a snapshot of a partially completed assembly. This type of recommendation could be useful both in the conceptual design stage and for part re-use recommendations during the modelling stage. To again enable quantitative evaluation we also frame this experiment as a classification problem. In particular, we embed a whole batch of 512  part strings, and for each document we remove one part at random and try to correctly identify that removed part from the rest of the batch of parts. We use only a batch of 512 candidates rather than the entire set of possible parts since the number of possible parts is very large. In fact it is an order of magnitude greater than the typical vocabulary size of a LLM.  We also wish to reduce the likelihood of including multiple very similar parts. While this simplifies the problem, all methods are tested in the same way, thus our comparisons hold.

For training (see \autoref{fig:missing_part_train}) we use the LLM embedding of the removed part directly as an auto-regressive target for a set encoder, which is given the remaining parts from the document as input. A Mean Squared Error (MSE) loss is used to minimize the distance between the predicted part embedding and the embedding of the target. For inference and evaluation (see \autoref{fig:missing_part_inference}) we mix the removed part string embedding into the embeddings for the rest of the part strings in the batch to form the target candidates. Then we use the cosine similarity to find the closest candidate to the predicted part embedding produced by the set encoder. A random classifier would be expected to achieve approximately 0.2\% accuracy on this task.

Due to the increased model size in this experiment, we train the set transformer on the full training set, however, the evaluation test set remains completely unseen by any of the language models.

\subsubsection{Document Name}

As a final experiment, we consider the task of predicting a document's name. This gives an indication of the extent to which the language model can effectively predict part-whole relationships. Recall that `documents' are loosely related to assemblies, as such in many cases it is feasible and realistic to attempt to identify the document's name from the parts contained within it. The ability to correctly classify a collection of parts can be applied directly to the automatic categorization of assemblies in a PLM system. Moreover, with an understanding of an assembly as a whole, CAD software can tailor quick access tools and documentation to suit the class of object being modelled and suggest relevant design standards to follow. We expect this task to be the hardest of the three, in that to solve it effectively on the unstructured and diverse data included in the ABC dataset would require significant `world knowledge'.

We frame this experiment as per the ``Missing Part" task, except for providing all non-default document parts as input and using the document names for the targets. Again using a batch size of 512, the expected performance of a random classifier would be 0.2\% accuracy.

\subsection{Fine-Tuning}
\label{sec:fine-tuning}

We note that we do not fine-tune the LLM separately for each task and fine-tune only once on a corpus generated from the ABC text data. To generate the corpus we use the following template string: ``\texttt{An assembly with name \{DOCUMENT\_NAME\} contains the following parts: \{PART\_1\}, \dots, \{PART\_N\}.\textbackslash n}". The pre-trained language model is then fine-tuned line-by-line for 3 epochs using all default settings with a script provided by Hugging Face \cite{huggingface}. This corpus also serves as a complete training corpus for our baseline models which require training (see \autoref{sec:baselines}).

\subsection{Baselines}
\label{sec:baselines}

For the most simple baselines we compare against a bag-of-words approach (BOW) using both frequency-based word vectors and term frequency–inverse document frequency (TF-IDF) word vectors. We also compare against the pre-trained TechNet word embeddings \cite{sarica2020}, which are mean pooled to provide embeddings for each part or document name. Finally, for a learning-based baseline trained entirely from scratch on the ABC text data, we compare against FastText \cite{bojanowski2017}, which is trained on the fine-tuning corpus generated from the full train set (see \autoref{sec:fine-tuning}). Training FastText from scratch allows us full control over the hyperparameters and thus we are able to sweep the embedding size over a comparable range to DistilBERT rather than adopting the pre-trained common crawl embeddings with a fixed embedding size of only 300 dimensions.

\subsection{Hyper-Parameters \& Implementation}
\label{sec:implementation}

All implementations are in pytorch with pytorch-lightning unless otherwise stated.

The MLP used in all ``Two Parts'' experiments has 100 hidden units, and is optimized using Adam with a learning rate of 0.001. For FastText we use the gensim implemtation and grid search embedding size in \{128, 256, 512, 768\}, window size in \{6, 8, 10\} and skipgram in \{TRUE, FALSE\}. See \autoref{sec:fine-tuning} for details on fine-tuning DistilBERT.

For the SetTransformer used in the ``Missing Part'' and ``Document Name'' experiments we grid search hidden dimensions in \{512, 768\}, number of in-directions in \{32, 64, 128\} and number of attention heads in \{4, 8\}. In all cases we found greater stability in training without layer norm, thus it is not used throughout. We train for a maximum of 200 epochs with early stopping on the validation loss with a patience of 40. Again we use Adam optimizer with a learning rate 0.001.

Any hyper-parameters for any method not detailed above are set to defaults.

\section{Results and Discussion}
\label{sec:results}

All results show the mean test accuracy with standard deviations after 5 trials on the held-out test set unless otherwise stated. In all cases `Random' refers to the expected value of a random classifier, `DistilBERT' refers to the pre-trained version, and `DistilBERT-FT' refers to the same pre-trained model with additional fine-tuning (see \autoref{sec:fine-tuning}) on the cleaned version of the ABC corpus which we contribute with this work.

%%%%%%%%%%%%%
% Two Parts %
%%%%%%%%%%%%%
\subsection{Two Parts}

\autoref{tab:two_parts} shows the mean test accuracies of the ``Two Parts" experiment. We observe that the traditional BOW approaches both score no better than random, indicating that simple statistics of repeated words are not sufficient to solve this task with any level of success. We also note that DistilBERT (without any fine-tuning or task-specific training) outperforms both the pre-trained TechNet embeddings (trained on mechanical and engineering design patents) and the FastText embeddings which are trained specifically on the ABC data with a subword tokenizer. This supports our hypothesis that general-purpose pre-trained LLMs contain valuable domain-specific mechanical engineering knowledge that can be leveraged in downstream tasks. Finally, we see that fine-tuning the language model on the ABC string data further improves the performance. While the gain is small, it has statistical significance (with $p$ value of less than $1\%$), and as such it shows that the string data contains additional information which the language model can learn to exploit. This may include domain-specific vocabulary, brand names, and part numbers, not present in the original Wikipedia and BookCorpus training set.   More training data of this kind might enable further improvement in results, leading to a better understanding of parts that co-occur in mechanical assemblies.

\begin{table}[h]
    \centering
    \caption{Mean test accuracy at predicting whether or not two parts are from the same document.}
    \begin{tabular}{lr}
    \toprule
    Language Model & Test Accuracy (\%) \\
    \midrule
    Random                  &      $50$ \\
    TF-IDF BOW              &      $50.1 \pm 2.0$ \\
    Frequency BOW           &      $49.8 \pm 1.8$ \\
    TechNet                 &      $61.6 \pm 4.1$ \\
    FastText                &      $66.6 \pm 1.9$ \\
    DistilBERT              &      $68.0 \pm 1.9$ \\
    DistilBERT-FT           &      $\mathbf{68.9 \pm 2.2}$ \\
    \bottomrule
    \end{tabular}
    \label{tab:two_parts}
\end{table}

Note, due to the close proximity of the top scores, we increased the number of trials to 100 in order to collect enough evidence to interpret any differences. For the other models with already much larger gaps, this was not necessary. Also, due to the complete failure of the BOW approaches, we do not consider these in further experiments.

%%%%%%%%%%%%%%%%
% MISSING PART %
%%%%%%%%%%%%%%%%
\subsection{Missing Part}

\autoref{tab:missing_part} shows the comparison of the mean test classification accuracies for each language model at identifying the correct missing part from a batch of 512 candidates.

\begin{table}[h]
    \centering
    \caption{Mean test accuracy for predicting a missing part from the remaining parts in a document.}
    \begin{tabular}{lr}
    \toprule      
    Language Model            &     Test Accuracy (\%) \\
    \midrule
    Random          &  $ 0.2$ \\
    TechNet         &  $ 1.8 \pm 0.5$ \\
    FastText        &  $20.8 \pm 1.3$ \\
    DistilBERT      &  $27.2 \pm 0.6$ \\
    DistilBERT-FT   &   $\mathbf{32.1 \pm 0.4}$ \\
    \bottomrule
    \end{tabular}
    \label{tab:missing_part}
\end{table}

Here TechNet performs very poorly, which is likely due to its limitation of whole-word tokens. Whereas in the ``Two Parts" task, the presence of out-of-vocab (OOV) tokens may have provided a clue to whether or not two parts came from the same assembly, in the more difficult setting here the semantics alone captured by TechNet word embeddings are not enough to predict the missing part with useful accuracy.

FastText (fully trained on the ABC text train set) performs much better, however the pre-trained LLM DistilBERT achieves better performance despite never having seen the ABC data before. Again this confirms our hypothesis that general-purpose LLMs contain knowledge useful for engineering and CAD-related tasks.

In this particular experiment we note that fine-tuning has a more substantial improvement in the predictions than in the ``Two Parts" case. Recall that the fine-tuning was not done directly on this task, but through a masked language modelling task on the generated corpus (see \autoref{sec:fine-tuning}). Thus, here we demonstrate that fine-tuning on the corpus provided with this work allows language models to generalize to new tasks in the domain of CAD and mechanical engineering.

%%%%%%%%%%%%%%%%%
% Document Name %
%%%%%%%%%%%%%%%%%
\subsection{Document Name}

\begin{table}[h]
    \centering
    \caption{Mean test accuracy for predicting a document name given the set of part names.}
    \begin{tabular}{lr}
    \toprule      
    Language Model            &     Test Accuracy (\%) \\
    \midrule
    Random          & $0.2$ \\
    TechNet         & $3.3 \pm 0.5$\\
    FastText        &   $11.3 \pm 0.6$ \\
    DistilBERT      &  $18.0 \pm 0.7$ \\
    DistilBERT-FT   & $\mathbf{19.3 \pm 1.9}$ \\
    \bottomrule
    \end{tabular}
    \label{tab:document_name}
\end{table}

\autoref{tab:document_name} shows the mean test accuracies for identifying the correct document name from a batch of 512 candidates given the document's parts as input.

With the exception of TechNet, for all language models this task presented the greatest challenge. This is largely expected, since the downstream model must go beyond recognizing common co-occurring parts to reasoning about part-whole relationships.

For TechNet, the improvement seen here (compared to the ``Missing Part'' experiment) is likely due to its OOV limitations in that we expect to find fewer OOV terms in the document names than the part names. Therefore, given a set of parts with only some OOV parts, predicting a missing part that is also OOV would not be possible, however predicting the name could be more achievable.

\begin{table}[htp]
\renewcommand{\arraystretch}{1.2}
    \caption{Missing part (part-part) predictions DistilBERT-FT and FastText both made correctly.}
    \centering
    \begin{tabularx}{\linewidth}{lX}
         \toprule
         Missing Part & Input Parts \\
         \midrule
         20.1134 30 r 30 & 20.1125 30 r 90, 20.1136 30 r 60 \\
motor plate & sensor plate, motor plate, slew bearing plate \\
joist 9 & concrete, house, roof, patio roof, joist 14, joist 13, joist 12, joist 11, joist 10, joist 8, joist 7, joist 6, joist 5, joist 4, joist 3, joist 2, joist 1, cross member, post 5, post 4, post 3, post 2, post 1 \\
front right wheel & back left wheel, front left wheel, back right wheel, back axel, front axel, large base, back wheels, front wheels, base \\
support leg - upper & support leg - lower, central leg \\
    pin-3 & base part, pin-2, pin-4, pin-5, link-1, pin-1, shaft \\
starboard outer & port outer, port inner, base, starboard inner, horn, frame \\
lower swivel bracket & upper swivel bracket, board mount plate \\
gear 2 & gear 4, gear 3, gear 1, construction \\
syringe piston & syringe body \\
         \bottomrule
    \end{tabularx}
    \label{tab:missingpartboth}
\end{table}

\begin{table}[htp]
\renewcommand{\arraystretch}{1.2}
    \caption{Missing part (part-part) predictions DistilBERT-FT made correctly where FastText failed.}
    \centering
    \begin{tabularx}{\linewidth}{lX}
         \toprule
         Missing Part & Input Parts \\
         \midrule
         bearing spacer & gt2 pulley 16, right motor, left motor, m5 washer, f695, nema17 \\
plate & switch, bar, bezel, base, mount \\
arm 1 & bent finger 3, bent finger 2, bent finger 1, finger 1, finger bent 1, body, shoulder 1, finger flat \\
handle & fork, wheel, clamp, grip, brake, deck \\
bearing plug & 52 mm wheel, mold base \\
rook & queen, castle, pawn, knight \\
tail & frontsc wings, t 25 aelius \\
base holder & clamp, umbrella pole \\
shelf & turntable, wall \\
lid & case \\
cover & holder, shell \\
base & dowel, gear \\
         \bottomrule
    \end{tabularx}
    \label{tab:missingpartbert}
\end{table}

Again the pre-trained DistilBERT outperforms both TechNet and FastText by a large margin, thus providing further support for our hypothesis that the ``world-knowledge" captured by such LLMs can be effectively leveraged for downstream CAD and mechanical engineering-related tasks.

%%%%%%%%%%%%%%%
% Qualitative %
%%%%%%%%%%%%%%%

\subsection{Qualitative Analysis}
\label{sec:qualitative}

Considering the ``Missing Part'' predictions made by each of the models, we observe many textual correlations in targets and input parts. It is very common in the dataset for parts from the same document to share words or sub-words which could give the models a way to cheat at identifying the correct candidate. We believe FastText (trained purely on the ABC data with a sub-word tokenizer)  provides a baseline for this, from which we can compare the language models and quantify any gains.

To illustrate this, \autoref{tab:missingpartboth} shows examples that both FastText and DistilBERT-FT were able to correctly predict. Here we see many common words or sub-words or a similar pattern of numbers. Such predictions could still be useful - i.e. knowing we likely need a ``front right wheel'' given a ``front left wheel'', etc. has value and the text provides a much simpler indicator of this than geometry alone. However, it is not possible to determine here whether the models really understand the relationships between the parts or are simply looking for common words/sub-words, which would not generalize well to a setting in which we have a much greater number of candidate parts.

\begin{table}[htp]
\renewcommand{\arraystretch}{1.2}
    \caption{Document name predictions that both DistilBERT-FT and FastText made correctly which include many non-semantic clues in the text.}
    \centering
    \begin{tabularx}{\linewidth}{lX}
         \toprule
         Document Name & Input Parts \\
         \midrule
         miniatures game box & miniatures game tray, cover, miniatures game box \\
t-25 helios base delta tail & tail, delta wings, t 25 helios base \\
ds spinner ball 15mm & skull pin 8.1mm, spinner top, spinner bottom, spinner roundcurved top, spinner roundcurved bottom, spinner ninja duo top, spinner ninja duo bottom, spinner curved top, spinner curved bottom, spinner dragon top, spinner dragon bottom, unicorn pin 8.1 r, ninjago spinner bottom, ninjago spinner top, spinner wing top, spinner wing bottom, spinner wing duo top, spinner wing duo bottom, ninjago pin, unicorn pin 8.1 l \\
t-25r sorin base ogoc wings & ogoc wings, left thruster, right thruster, sorin base \\
bluetooth speaker & speaker small, speaker large \\
ctc e3d titan mount & pcb mount, sensor mount, bearing mount, e3d blower, carriage, titan mount \\
din 912 m3 screw & din 912 m3x14, din 912 m3x12, din 912 m3x10, din 912 m3x8, din 912 m3x6, din 912 m3x4 \\
chair tm8 blocks2.step & chair tm8 blocks2, chair tm8 blocks2, chair tm8 blocks2, chair tm8 blocks2, chair tm8 blocks2, chair tm8 blocks2, chair tm8 blocks2, chair tm8 blocks2, chair tm8 blocks2, chair tm8 blocks2, chair tm8 blocks2, chair tm8 blocks2, chair tm8 blocks2, chair tm8 blocks2, chair tm8 blocks2, chair tm8 blocks2, chair tm8 blocks2, chair tm8 blocks2 \\
2020 clip lock mount for led lamp & lock, mount \\
t-25 aelius base starfighter tail fixing & tail, starfighter wings, t 25 aelius \\
         \bottomrule
    \end{tabularx}
    \label{tab:docnameboth}
\end{table}

\begin{table}[htp]
\renewcommand{\arraystretch}{1.2}
    \caption{Semantically-based document name predictions DistilBERT-FT got correct that FastText did not.}
    \centering
    \begin{tabularx}{\linewidth}{lX}
    \toprule
    Document Name & Input Parts \\
    \midrule
    suspension & tire 2, wheel 2, upright 2, tire 1, wheel 1, lower a-arm 2, upper a-arm 2, upper a-arm 1, upright 1, lower a-arm mount, lower a-arm 1 \\
service counter & top surface left b, bottom drawer unit, side skirt left, front skirt full, side skirt right, work table surface b, upper storage b-right, 4 inch top writing surface, top surface right b, middle deck left panel 2, upper storage b-left, upper side skirt right, top trim, top front skirt b \\
chess-game & bishop, knight, base, queen, pawn-top, pawn-base, king, rook \\
si15024 org2 radio-controlled helicopter & tail propeller screw, propeller screw, tail rotor gear base axis, tail rotor gear base gear, base, skid 2, skid 1, tail fin screw, tail fin, tail propeller 2, tail propeller 1, tail rotor base, tail rotor axis, tail rotor gear, tail rotor gear t, tail motor base gear t, tail motor base p, tail motor axis, motor, motor gear, propeller pole gear, motor pole, propeller2, propeller1, propeller base, canopy, pole \\
gimbal & right servo, right gear, left servo, center t, top gear, left gear, frame \\
bullet feeder & nose guide spacer, collator, nose guide, flip ramp, collator side wall, tube attachment, base mounting bracket, motor \\
ball bearing & inner ring, outer ring \\
brushless rc car starter document & tg-3 receiver, m3x14 bolt, m3x12 bolt, 625 bearing, shaft, upper face, main casing, lower face, motor mount, plastic input, heatsink, base pcb, plastic base, m3x25 bolt, servo mount screw, pla, m3 nut, m5x40 bolt standard head, m3x8 bolt, m3x20 bolt, m3x35 bolt, m3x16 bolt, base plate, 9g micro servo, m5 nut, rear drive gear, rear axel, 500mah 3s lipo, m3x10 bolt, m3x30 bolt \\
robot arm & pot gear, upper clevis, lower clevis, rod knuckle, elbow idler, tube socket, lifter arm rotor, base tube rotor, riser base, fore clevis, rear clevis, base support r, base support l, can, gripper frame, actuator arm, servo shaft, servo, right gripper, left gripper, right frame, base bottom, left frame, main gear \\
robot & hbridge holder, arduino holder, shaft, fr wheel, wheel holder, arduino uno, h-bridge, battery, sensor-holder, pin, sensor, base, bracket, motor, wheel, pin \\
\bottomrule
    \end{tabularx}
    \label{tab:docnamebert}
\end{table}

\autoref{tab:missingpartbert} shows examples that DistilBERT-FT was able to correctly predict where FastText failed. In contrast, here we see examples where the semantic understanding of part-part relations is necessary in order to make the correct prediction, and confirm that the DistilBERT-FT model is able to make predictions based on more than just text matching.

We see a similar situation looking at the ``Document Name" predictions. Looking at examples both FastText and DistilBERT-FT were able to correctly predict (\autoref{tab:docnameboth}), we again see many common words in the document names compared with the input parts. However, when considering only examples that DistilBERT-FT predicted correctly where FastText failed (\autoref{tab:docnamebert}) we again see examples that could not be correctly predicted without part-whole semantic knowledge.

%%%%%%%%%%%%%%%
% Limitations %
%%%%%%%%%%%%%%%

\subsection{Limitations}
\label{sec:limitations}

The performance of any LLM (or in fact any model) operating on text alone will be limited by how well this text describes the geometry or function of the part(s) it is associated with. As shown in \autoref{tab:example_names} and \ref{tab:example_parts}, the ABC dataset has many ambiguous strings that bear little or no relation to the geometry or function of the parts they describe, and we expect that in industrial engineering data we may find a similar situation. While we leave multi-modal extension to further work, we hypothesize that the inclusion of geometry will enable models to overcome this inherent limitation of LLMs and draw more value than from each modality alone, thus offering a means to better represent and reason with part affordances. We therefore consider multi-modal investigation with geometry to be highly justified.

Additionally, we note our finding that the text data contained in CAD document parts is often highly correlated in ways that make it easy for language-based models to find trivial solutions, i.e. predicting a missing ``pin 3" when presented with ``pin 1", ``pin 2" and ``pin 4". As discussed in \autoref{sec:qualitative} we believe using a FastText baseline enabled us to better understand any semantic gains over these trivial solutions and we note the importance of such considerations for any future work using CAD text data.

Given our desire to make quantitative evaluations, we opted for a classification style setup when predicting a missing part or document name. While this makes it easier to compare different methods and draw meaningful conclusions, it limits our approach to settings where a set of plausible candidate targets can be provided. For many use cases this may be appropriate, e.g. recommendations of  fastener or other commercially available parts, however, extensions of this work may seek to provide generated open-ended natural language outputs for the predictions.

\section{Conclusion}
In this work we demonstrate that pre-trained LLMs contain substantial mechanical engineering knowledge, outperforming numerous benchmarks on a number of self-supervised tasks by a significant margin. We show that traditional statistical word-count approaches are insufficient to handle the noisy natural language contained in CAD files, and surprisingly, we observe that a pre-trained language model performs better on these tasks than TechNet, which is specifically trained   on mechanical engineering patent data, and in some cases improves by an order of magnitude.    

We also demonstrate that fine-tuning the model on CAD data, using a standard masked token predication language modelling task, leads to improvement in performance on all of the downstream self-supervised tasks tested.   This confirms our hypothesis that unstructured natural language text information inside CAD files can be utilized to improve language models understanding of mechanical engineering concepts.   

We publicly release the cleaned text data corpus extracted from the ABC dataset, the first CAD text corpus of its kind, along with all of our code in the hope that this will inspire future research and can be used for bench-marking future tasks.

In addition to the contributions discussed above, this work also identifies key limitations to using LLMs with CAD text data alone. For example, the large number of trivial solutions enabled by highly correlated text in part names within documents, for which we propose that FastText with a subword tokenizer can provide a basleine. We also provide an insight into the proportion of part and document names within the dataset that actually provide useful semantic or geometric information creating a strong motivation for multi-modal approaches including geometry.

In future work, we plan to investigate how embeddings from multi-modal text-CAD training can allow better semantic clustering of geometries into groups with similar affordance. Moreover, the findings of our work could also be leveraged to improve and support generative workflows in CAD software.

%%%%%%%%%%%%%%%%%%%%%%%%%%%%%%%%%%%%%%%%%%%%%%%%%%%%%%%%%%%%%%%%%%%%%%
\begin{acknowledgment}
Thanks go to the authors of TechNet, in particular Serhad Sarica and Jianxi Luo, for their support with the TechNet pre-processing pipeline, and to Pradeep Kumar Jayaraman for his advice and guidance on transformers.
\end{acknowledgment}

%%%%%%%%%%%%%%%%%%%%%%%%%%%%%%%%%%%%%%%%%%%%%%%%%%%%%%%%%%%%%%%%%%%%%%
% The bibliography is stored in an external database file
% in the BibTeX format (file_name.bib).  The bibliography is
% created by the following command and it will appear in this
% position in the document. You may, of course, create your
% own bibliography by using thebibliography environment as in
%
% \begin{thebibliography}{12}
% ...
% \bibitem{itemreference} D. E. Knudsen.
% {\em 1966 World Bnus Almanac.}
% {Permafrost Press, Novosibirsk.}
% ...
% \end{thebibliography}

% Here's where you specify the bibliography style file.
% The full file name for the bibliography style file
% used for an ASME paper is asmems4.bst.
\bibliographystyle{asmems4}

% Here's where you specify the bibliography database file.
% The full file name of the bibliography database for this
% article is asme2e.bib. The name for your database is up
% to you.
\bibliography{asme2e}

\begin{thebibliography}{10}

\bibitem{Bespalov2005}
Bespalov, D., Ip, C.~Y., Regli, W.~C., and Shaffer, J., 2005.
\newblock ``Benchmarking cad search techniques''.
\newblock In Proceedings of the 2005 ACM Symposium on Solid and Physical
  Modeling, SPM '05, Association for Computing Machinery, p.~275–286.

\bibitem{kwon2022}
Kwon, E., Huang, F., and Goucher-Lambert, K., 2022.
\newblock ``Enabling multi-modal search for inspirational design stimuli using
  deep learning''.
\newblock {\em Artificial Intelligence for Engineering Design, Analysis and
  Manufacturing, {\bf 36}}, p.~e22.

\bibitem{korbi2022cad}
Korbi, A., Tlija, M., and Louhichi, B., 2022.
\newblock ``A cad model for the tolerancing of mechanical assemblies
  considering non-rigid joints between parts with defects''.
\newblock {\em Proceedings of the Institution of Mechanical Engineers, Part B:
  Journal of Engineering Manufacture, {\bf 236}}(3), pp.~219--232.

\bibitem{jones2021automate}
Jones, B., Hildreth, D., Chen, D., Baran, I., Kim, V.~G., and Schulz, A., 2021.
\newblock ``Automate: A dataset and learning approach for automatic mating of
  cad assemblies''.
\newblock {\em ACM Transactions on Graphics (TOG), {\bf 40}}(6), pp.~1--18.

\bibitem{willis2022joinable}
Willis, K.~D., Jayaraman, P.~K., Chu, H., Tian, Y., Li, Y., Grandi, D., Sanghi,
  A., Tran, L., Lambourne, J.~G., Solar-Lezama, A., and Matusik, W., 2022.
\newblock ``Joinable: Learning bottom-up assembly of parametric cad joints''.
\newblock In Proceedings of the IEEE/CVF Conference on Computer Vision and
  Pattern Recognition (CVPR), pp.~15849--15860.

\bibitem{Shi2017}
Shi, F., Chen, L., Han, J., and Childs, P., 2017.
\newblock ``A data-driven text mining and self-learning semantic network
  analysis for design knowledge retrieval''.
\newblock {\em Journal of Mechanical Design, {\bf 139}}, 08.

\bibitem{sarica2020}
Sarica, S., Luo, J., and Wood, K.~L., 2020.
\newblock ``{{TechNet}}: {{Technology Semantic Network Based}} on {{Patent
  Data}}''.
\newblock {\em Expert Systems with Applications, {\bf 142}}, Mar., p.~112995.

\bibitem{Feng2020}
Feng, Y., Zhao, Y., Zheng, H., Li, Z., and Tan, J., 2020.
\newblock ``Data-driven product design toward intelligent manufacturing: A
  review''.
\newblock {\em International Journal of Advanced Robotic Systems, {\bf 17}}(2),
  p.~1729881420911257.

\bibitem{Schinko2017}
Schinko., C., Vosgien., T., Prante., T., Schreck., T., and Ullrich., T., 2017.
\newblock ``Search \& retrieval in cad databases - a user-centric
  state-of-the-art overview''.
\newblock In Proceedings of the 12th International Joint Conference on Computer
  Vision, Imaging and Computer Graphics Theory and Applications - GRAPP,
  (VISIGRAPP 2017), INSTICC, SciTePress, pp.~306--313.

\bibitem{tan2020neural}
Tan, Z., Wang, S., Yang, Z., Chen, G., Huang, X., Sun, M., and Liu, Y., 2020.
\newblock ``Neural machine translation: A review of methods, resources, and
  tools''.
\newblock {\em AI Open, {\bf 1}}, pp.~5--21.

\bibitem{raffel2020exploring}
Raffel, C., Shazeer, N., Roberts, A., Lee, K., Narang, S., Matena, M., Zhou,
  Y., Li, W., Liu, P.~J., et~al., 2020.
\newblock ``Exploring the limits of transfer learning with a unified
  text-to-text transformer.''.
\newblock {\em J. Mach. Learn. Res., {\bf 21}}(140), pp.~1--67.

\bibitem{brown2020language}
Brown, T., Mann, B., Ryder, N., Subbiah, M., Kaplan, J.~D., Dhariwal, P.,
  Neelakantan, A., Shyam, P., Sastry, G., Askell, A., et~al., 2020.
\newblock ``Language models are few-shot learners''.
\newblock {\em Advances in neural information processing systems, {\bf 33}},
  pp.~1877--1901.

\bibitem{wiki}
Website, 2019.
\newblock English wikipedia.
\newblock \url{https://en.wikipedia.org/wiki/English\_Wikipedia}.

\bibitem{Zhu_2015_ICCV}
Zhu, Y., Kiros, R., Zemel, R., Salakhutdinov, R., Urtasun, R., Torralba, A.,
  and Fidler, S., 2015.
\newblock ``Aligning books and movies: Towards story-like visual explanations
  by watching movies and reading books''.
\newblock In Proceedings of the IEEE International Conference on Computer
  Vision (ICCV).

\bibitem{Mehmood2017CommonCrawl}
Mehmood, M.~A., Shafiq, H.~M., and Waheed, A., 2017.
\newblock ``Understanding regional context of world wide web using common crawl
  corpus''.
\newblock In 2017 IEEE 13th Malaysia International Conference on Communications
  (MICC), pp.~164--169.

\bibitem{koch2019abc}
Koch, S., Matveev, A., Jiang, Z., Williams, F., Artemov, A., Burnaev, E.,
  Alexa, M., Zorin, D., and Panozzo, D., 2019.
\newblock ``Abc: A big cad model dataset for geometric deep learning''.
\newblock In Proceedings of the IEEE Conference on Computer Vision and Pattern
  Recognition, pp.~9593--9603.

\bibitem{vaswani2017attention}
Vaswani, A., Shazeer, N., Parmar, N., Uszkoreit, J., Jones, L., Gomez, A.~N.,
  Kaiser, {\L}., and Polosukhin, I., 2017.
\newblock ``Attention is all you need''.
\newblock {\em Advances in neural information processing systems, {\bf 30}}.

\bibitem{ZHANG2021205}
Zhang, C., Lai, Y., Feng, Y., and Zhao, D., 2021.
\newblock ``A review of deep learning in question answering over knowledge
  bases''.
\newblock {\em AI Open, {\bf 2}}, pp.~205--215.

\bibitem{Radford2018}
Radford, A., Wu, J., Child, R., Luan, D., Amodei, D., and Sutskever, I., 2018.
\newblock ``Language models are unsupervised multitask learners''.

\bibitem{devlin2019}
Devlin, J., Chang, M., Lee, K., and Toutanova, K., 2019.
\newblock ``{BERT:} pre-training of deep bidirectional transformers for
  language understanding''.
\newblock In Proceedings of the 2019 Conference of the North American Chapter
  of the Association for Computational Linguistics: Human Language
  Technologies, {NAACL-HLT} 2019, Minneapolis, MN, USA, June 2-7, 2019, Volume
  1 (Long and Short Papers), J.~Burstein, C.~Doran, and T.~Solorio, eds.,
  Association for Computational Linguistics, pp.~4171--4186.

\bibitem{sanh2020}
Sanh, V., Debut, L., Chaumond, J., and Wolf, T., 2020.
\newblock {{DistilBERT}}, a distilled version of {{BERT}}: Smaller, faster,
  cheaper and lighter, Feb.

\bibitem{rogers2020}
Rogers, A., Kovaleva, O., and Rumshisky, A., 2020.
\newblock ``A primer in {BERT}ology: What we know about how {BERT} works''.
\newblock {\em Transactions of the Association for Computational Linguistics,
  {\bf 8}}, pp.~842--866.

\bibitem{lee2019}
Lee, J., Lee, Y., Kim, J., Kosiorek, A., Choi, S., and Teh, Y.~W., 2019.
\newblock ``Set transformer: A framework for attention-based
  permutation-invariant neural networks''.
\newblock In Proceedings of the 36th International Conference on Machine
  Learning, pp.~3744--3753.

\bibitem{radford2021}
Radford, A., Kim, J.~W., Hallacy, C., Ramesh, A., Goh, G., Agarwal, S., Sastry,
  G., Askell, A., Mishkin, P., Clark, J., Krueger, G., and Sutskever, I., 2021.
\newblock Learning {{Transferable Visual Models From Natural Language
  Supervision}}, Feb.

\bibitem{rombach2022}
Rombach, R., Blattmann, A., Lorenz, D., Esser, P., and Ommer, B., 2022.
\newblock ``High-resolution image synthesis with latent diffusion models''.
\newblock In Proceedings of the IEEE/CVF Conference on Computer Vision and
  Pattern Recognition (CVPR), pp.~10684--10695.

\bibitem{Tangelder2004}
Tangelder, J., and Veltkamp, R., 2004.
\newblock ``A survey of content based 3d shape retrieval methods''.
\newblock In Proceedings Shape Modeling Applications, 2004., pp.~145--156.

\bibitem{Fisher2010}
Fisher, M., and Hanrahan, P., 2010.
\newblock ``Context-based search for 3d models''.
\newblock {\em ACM Trans. Graph., {\bf 29}}(6), dec.

\bibitem{Funkhouser2003}
Funkhouser, T., Min, P., Kazhdan, M., Chen, J., Halderman, A., Dobkin, D., and
  Jacobs, D., 2003.
\newblock ``A search engine for 3d models''.
\newblock {\em ACM Trans. Graph., {\bf 22}}(1), jan, p.~83–105.

\bibitem{Salton1991}
Salton, G., 1991.
\newblock ``The smart document retrieval project''.
\newblock In Proceedings of the 14th Annual International ACM SIGIR Conference
  on Research and Development in Information Retrieval, SIGIR '91, Association
  for Computing Machinery, p.~356–358.

\bibitem{rocchio71relevance}
Rocchio, J.~J., 1971.
\newblock ``Relevance feedback in information retrieval''.
\newblock In {\em The Smart retrieval system - experiments in automatic
  document processing}, G.~Salton, ed. Englewood Cliffs, NJ: Prentice-Hall,
  pp.~313--323.

\bibitem{Yi2017}
Yi, L., Guibas, L., Hertzmann, A., Kim, V.~G., Su, H., and Yumer, E., 2017.
\newblock ``Learning hierarchical shape segmentation and labeling from online
  repositories''.
\newblock {\em SIGGRAPH}.

\bibitem{chang2015shapenet}
Chang, A.~X., Funkhouser, T., Guibas, L., Hanrahan, P., Huang, Q., Li, Z.,
  Savarese, S., Savva, M., Song, S., Su, H., et~al., 2015.
\newblock ``Shapenet: An information-rich 3d model repository''.
\newblock {\em arXiv preprint arXiv:1512.03012}.

\bibitem{chen2018text2shape}
Chen, K., Choy, C., Savva, M., Chang, A., Funkhouser, T., and Savarese, S.,
  2019.
\newblock ``Text2shape: Generating shapes from natural language by learning
  joint embeddings''.
\newblock In Computer Vision - ACCV 2018 - 14th Asian Conference on Computer
  Vision, Revised Selected Papers, H.~Li, C.~Jawahar, K.~Schindler, and
  G.~Mori, eds., Lecture Notes in Computer Science (including subseries Lecture
  Notes in Artificial Intelligence and Lecture Notes in Bioinformatics),
  Springer Verlag, pp.~100--116.
\newblock Funding Information: Acknowledgments. This material is based upon
  work supported by the National Science Foundation Graduate Research
  Fellowship Program under Grant No. DGE – 1147470. Any opinions, findings,
  and conclusions or recommendations expressed in this material are those of
  the author(s) and do not necessarily reflect the views of the National
  Science Foundation. This work is supported by Google, Intel, and with the
  support of the Technical University of Munich–Institute for Advanced Study,
  funded by the German Excellence Initiative and the European Union Seventh
  Framework Programme under grant agreement no. 291763. Publisher Copyright:
  {\textcopyright} 2019, Springer Nature Switzerland AG.; 14th Asian Conference
  on Computer Vision, ACCV 2018 ; Conference date: 02-12-2018 Through
  06-12-2018.

\bibitem{Han2019}
Han, Z., Shang, M., Wang, X., Liu, Y.-S., and Zwicker, M., 2019.
\newblock ``Y2seq2seq: Cross-modal representation learning for 3d shape and
  text by joint reconstruction and prediction of view and word sequences''.
\newblock {\em Proceedings of the AAAI Conference on Artificial Intelligence,
  {\bf 33}}(01), Jul., pp.~126--133.

\bibitem{ramesh2021zero}
Ramesh, A., Pavlov, M., Goh, G., Gray, S., Voss, C., Radford, A., Chen, M., and
  Sutskever, I., 2021.
\newblock ``Zero-shot text-to-image generation''.
\newblock In International Conference on Machine Learning, PMLR,
  pp.~8821--8831.

\bibitem{dalle}
Dayma, B., Patil, S., Cuenca, P., Saifullah, K., Abraham, T., Le~Khac, P.,
  Melas, L., and Ghosh, R., 2021.
\newblock Dalle mini, 7.

\bibitem{sanghi2022clip}
Sanghi, A., Chu, H., Lambourne, J.~G., Wang, Y., Cheng, C.-Y., Fumero, M., and
  Malekshan, K.~R., 2022.
\newblock ``Clip-forge: Towards zero-shot text-to-shape generation''.
\newblock In Proceedings of the IEEE/CVF Conference on Computer Vision and
  Pattern Recognition, pp.~18603--18613.

\bibitem{khalid2022clip}
Khalid, N., Xie, T., Belilovsky, E., and Popa, T., 2022.
\newblock ``Clip-mesh: Generating textured meshes from text using pretrained
  image-text models''.
\newblock {\em ACM Transactions on Graphics (TOG), Proc. SIGGRAPH Asia}.

\bibitem{sanghi2022textcraft}
Sanghi, A., Fu, R., Liu, V., Willis, K., Shayani, H., Khasahmadi, A.~H.,
  Sridhar, S., and Ritchie, D., 2022.
\newblock ``Textcraft: Zero-shot generation of high-fidelity and diverse shapes
  from text''.
\newblock {\em arXiv preprint arXiv:2211.01427}.

\bibitem{Schlachter2022}
Schlachter, K., Ahlbrand, B., Wang, Z., Perlin, K., and Ortenzi, V., 2022.
\newblock ``Zero-shot multi-modal artist-controlled retrieval and exploration
  of 3d object sets''.
\newblock In SIGGRAPH Asia 2022 Technical Communications, SA '22, Association
  for Computing Machinery.

\bibitem{Szykman2000}
Szykman, S., Sriram, R.~D., Bochenek, C., Racz, J.~W., and Senfaute, J., 2000.
\newblock ``{Design Repositories: Engineering Design's New Knowledge Base}''.
\newblock {\em IEEE Intelligent Systems and Their Applications, {\bf 15}}(3),
  pp.~48--55.

\bibitem{Bohm2004}
Bohm, M.~R., and Stone, R.~B., 2004.
\newblock ``{Product design support: Exploring a design repository system}''.
\newblock {\em American Society of Mechanical Engineers, Computers and
  Information in Engineering Division, CED}, pp.~55--65.

\bibitem{Bohm2008}
Bohm, M.~R., Stone, R.~B., Simpson, T.~W., and Steva, E.~D., 2008.
\newblock ``{Introduction of a data schema to support a design repository}''.
\newblock {\em CAD Computer Aided Design, {\bf 40}}(7), pp.~801--811.

\bibitem{Phelan2014}
Phelan, K., Wilson, C., and Summers, J.~D., 2014.
\newblock ``{Development of a design for manufacturing rules database for use
  in instruction of dfm practices}''.
\newblock In Proceedings of the ASME Design Engineering Technical Conference,
  Vol.~1A, pp.~1--7.

\bibitem{Bharadwaj2019}
Bharadwaj, A., Xu, Y., Angrish, A., Chen, Y., and Starly, B., 2019.
\newblock ``{Development of a pilot manufacturing cyberinfrastructure with an
  information rich mechanical cad 3D model repository}''.
\newblock {\em ASME 2019 14th International Manufacturing Science and
  Engineering Conference, MSEC 2019, {\bf 1}}, pp.~1--8.

\bibitem{Kurtoglu2005}
Kurtoglu, T., Campbell, M.~I., Bryant, C.~R., Stone, R.~B., and Mcadams, D.~A.,
  2005.
\newblock ``{Deriving a Component Basis for Computational Functional
  Synthesis}''.
\newblock In ICED 05: 15th International Conference on Engineering Design:
  Engineering Design and the Global Economy.

\bibitem{Hirtz2002}
Hirtz, J., Stone, R.~B., McAdams, D.~A., Szykman, S., and Wood, K.~L., 2002.
\newblock ``{A functional basis for engineering design: Reconciling and
  evolving previous efforts}''.
\newblock {\em Research in Engineering Design - Theory, Applications, and
  Concurrent Engineering, {\bf 13}}(2), pp.~65--82.

\bibitem{Cheong2011}
Cheong, H., Chiu, I., Shu, L.~H., Stone, R.~B., and McAdams, D.~A., 2011.
\newblock ``{Biologically meaningful keywords for functional terms of the
  functional basis}''.
\newblock {\em Journal of Mechanical Design, Transactions of the ASME, {\bf
  133}}(2), pp.~1--11.

\bibitem{Ferrero2020a}
Ferrero, V., 2020.
\newblock {PyDamp: Python-based Data Addition and Management of PSQL
  databases}.

\bibitem{sciencedirect}
Website, 2023.
\newblock Sciencedirect.
\newblock \url{https://www.sciencedirect.com/}.

\bibitem{Miller1995}
Miller, G.~A., 1995.
\newblock ``{WordNet}''.
\newblock {\em Communications of the ACM, {\bf 38}}(11), nov, pp.~39--41.

\bibitem{Liu2004}
Liu, H., and Singh, P., 2004.
\newblock ``{ConceptNet - a practical commonsense reasoning tool-kit}''.
\newblock {\em BT Technology Journal, {\bf 22}}(4), pp.~211--226.

\bibitem{carlson2010}
Carlson, A., Betteridge, J., Kisiel, B., Settles, B., Jr., E. R.~H., and
  Mitchell, T.~M., 2010.
\newblock ``Toward an architecture for never-ending language learning''.
\newblock In Proceedings of the Twenty-Fourth Conference on Artificial
  Intelligence (AAAI 2010).

\bibitem{Mikolov2013word2vec}
Mikolov, T., Sutskever, I., Chen, K., Corrado, G., and Dean, J., 2013.
\newblock ``Distributed representations of words and phrases and their
  compositionality''.
\newblock In Proceedings of the 26th International Conference on Neural
  Information Processing Systems - Volume 2, NIPS'13, Curran Associates Inc.,
  p.~3111–3119.

\bibitem{bian2022material}
Bian, S., Grandi, D., Hassani, K., Sadler, E., Borijin, B., Fernandes, A.,
  Wang, A., Lu, T., Otis, R., Ho, N., et~al., 2022.
\newblock ``Material prediction for design automation using graph
  representation learning''.
\newblock In International Design Engineering Technical Conferences and
  Computers and Information in Engineering Conference, Vol.~86229, American
  Society of Mechanical Engineers, p.~V03AT03A001.

\bibitem{huggingface}
Website, 2023.
\newblock Hugging face.
\newblock \url{https://huggingface.co/}.

\bibitem{birdsteven2009}
Bird, S., Loper, E., and Klein, E., 2009.
\newblock {\em Natural {{Language Processing}} with {{Python}}}.
\newblock {O'Reilly Media Inc.}

\bibitem{bojanowski2017}
Bojanowski, P., Grave, E., Joulin, A., and Mikolov, T., 2017.
\newblock ``{Enriching Word Vectors with Subword Information}''.
\newblock {\em Transactions of the Association for Computational Linguistics,
  {\bf 5}}, 06, pp.~135--146.

\end{thebibliography}

%%%%%%%%%%%%%%%%%%%%%%%%%%%%%%%%%%%%%%%%%%%%%%%%%%%%%%%%%%%%%%%%%%%%%%
% The appendix
%\input{sections/appendix/appendix}

\end{document}